\documentclass[11pt]{amsart}

\usepackage[subpreambles=false]{standalone}
\usepackage[dvipsnames]{xcolor}
\usepackage[utf8]{inputenc}
\usepackage[english]{babel}
\usepackage{multicol}
\usepackage{graphicx}
\usepackage[margin=2.5cm]{geometry}
\usepackage{comment}

\usepackage{amsmath,amsfonts,amssymb}

\usepackage{fnpct}
\PassOptionsToPackage{hyphens}{url}
\usepackage{hyperref}
\hypersetup{
breaklinks=true,
}
\usepackage[anythingbreaks]{breakurl}

\definecolor{-yellow}{HTML}{FFCE00}
\definecolor{-yellow-darker}{HTML}{dbaf00}
\definecolor{-green}{HTML}{007849}
\definecolor{-blue}{HTML}{0375B4}
\definecolor{-blue-darker}{HTML}{035582}

\usepackage[authoryear,sort&compress]{natbib}

\title{Machine Learning and Information Theory Concepts towards an AI Mathematician}
\author{Yoshua Bengio and Esmeralda S. Whitammer}
\begin{document}

\maketitle
\begin{abstract}
The current state-of-the-art in artificial intelligence is impressive, especially in terms of mastery of language, but not so much in terms of mathematical reasoning. What could be missing? Can we learn something useful about that gap from how the brains of mathematicians go about their craft? This essay builds on the idea that current deep learning mostly succeeds at system 1 abilities -- which correspond to our intuition and habitual behaviors -- but still lacks something important regarding system 2 abilities -- which include reasoning and robust uncertainty estimation. It takes an information-theoretical posture to ask questions about what constitutes an interesting mathematical statement, which could guide future work in crafting an AI mathematician. The focus is not on proving a given theorem but on discovering new and interesting \emph{conjectures}. The central hypothesis is that a desirable body of theorems better summarizes the set of all provable statements, for example by having a small description length while at the same time being close (in terms of number of derivation steps) to many provable statements.
\end{abstract}

\section{Context}
\label{sec:intro}

Research in artificial intelligence has advanced in tremendous and even scary~\citep{russell2019human} ways in the last couple of decades, fueled by research on deep learning~\citep{bengio2021deep}. Modern generative AI systems are particularly impressive because they show an ability to create realistic new \emph{content}: images, sounds, and text, for example. These systems are deep neural networks: functions used to represent a conditional distribution $P(Y|X)$ over values of a random variable $Y$, representing content, given the value of a random variable $X$, representing context or input. The input $X$ could, for example, be a natural language query while the output $Y$ could be a text, an image, or a video. Many of these systems are effectively regression models fit to a set of training examples $(x,y)$. The distinguishing feature of deep learning is that the number of degrees of freedom in the regression is very large, sometimes billions or trillions of parameters. A generative neural network can sample $Y$ given $X$ from the learned distribution and then use it as part of the next input. In the case of large language models (LLMs), $Y$ is the next word that follows textual context $X$, and the iterated applications of such a generator can be used to form a dialogue with LLMs like ChatGPT~\citep{chatgpt,gpt4}. A pre-training phase~\citep{bengio2021deep} may exploit larger datasets where we observe only inputs $x$ or a combination of unlabeled ($x$ only) and labeled (($x,y$) pairs) data, or where the training objective is less targeted to a specific task and more towards capturing the statistical structure in some or all of the random variables of interest.

What is surprising is how successful such approaches have been when the number of training examples and parameters are both scaled up, in spite of the fairly simple principles involved. For example, GPT-4~\citep{gpt4} is believed to have about a trillion parameters trained with more than a trillion examples (of predicting the next word $Y$ in the context of a piece of text $X$). In addition to raw scale of the models and datasets, some of the ingredients of that success are algorithmic: training the whole thing by stochastic gradient descent~\citep{bottou2010large} (i.e., updating the parameters in the direction of improving the training objective for just a random subset of examples, and repeating this many many times), the use of learned vectorial representations for each word~\citep{Bengio-et-al-NIPS2000} or more generally for symbols~\citep{hinton1987using}, the ``deep learning" idea~\citep{LeCun-et-al-Nature2015} of stacking many layers of jointly trained non-linear transformations, the learnable non-linear computations inspired by human attention~\citep{bahdanau2014neural} mechanisms, focusing computations on a few elements of a set at a time (by giving them a larger weight), or the extension of this attention mechanism to attend to multiple elements in parallel and stack these attention layers, also known as the popular Transformer~\citep{vaswani2017attention} architecture.

We now have AI systems that seem to master language. What is missing in order to achieve the kind of mathematical thinking abilities that many of us enjoy?
Could we build an AI system that could one day surpass our mathematical abilities? These are the questions motivating the discussions below, with an emphasis on connecting the current state-of-the-art in deep learning to human-level reasoning and mathematical skills.

The main question we aim at is the following: what could serve as a training objective for mathematical discovery? We propose an information-theoretic answer: one would prefer a library of mathematical statements that is small while making it easy to prove many provable statements. We also consider the exploratory process of generating conjectures based on such an objective and then using a goal-conditioned machine learning approach to try proving them by generating subgoals corresponding to lemmas.

This paper can be read as an initial answer from a machine learning expert to a graduate student who would like to explore machine learning methods to build an AI mathematician, one that could not only search for proofs or disproofs of conjectures but also generate interesting new conjectures.

\section{How does our brain think?}

When analyzing the strengths and weaknesses of deep learning systems, we have hypothesized~\citep{bengio2017consciousness,goyal2022inductive} that they are fairly good at capturing what psychologists~\citep{Kahneman-2011} call system 1 or intuitive abilities, while they are weaker at system 2 or deliberate reasoning abilities. System 1 abilities are associated with intuition and habitual behavior, what we can achieve without having to think about it. System 2 abilities are associated with reasoning, explicit thinking, deliberate planning, imagining scenarios and counterfactuals, etc. They allow us to filter the ideas, thoughts and impulses coming out of our intuitive generative machinery, spotting and fixing incoherent bits, making it possible to construct and manipulate complex abstractions in our mind. The work of a mathematician, or of most researchers for that matter, clearly requires a combination of both: generative intuition suggests paths to investigate and a rational understanding of what is mathematically coherent allows us to select and think through those that are most promising because they are interesting and more likely to be consistent with everything else we know. It is also plausible that the result of this coherence checking provides a training signal for further improving the mathematical intuition machine that generates new ideas and conjectures.

Inductive biases are structural constraints in learning systems that cause the learner to favour some hypotheses over others. Machine learning researchers can bring inductive biases, often motivated by biological systems, to reduce the scope of possible learned systems to those that are consistent with their preferences and known structure in the data.
For example, a commonly used inductive bias in machine learning is that the learner should prefer smoother functions, or in the case of categorizing objects in images, ones that are somewhat invariant to small geometric transformations of the input.
~\citet{bengio2017consciousness,goyal2022inductive} hypothesized that reasoning abilities have evolved thanks to a set of inductive biases such as the presence of the working memory bottleneck (we can only reliably hold around a handful of items in our conscious mind). Why would we have such a strong limitation? Maybe because it forces us to compress information (from the raw observed data) through the process of creating and exploiting abstractions. This hypothesis is consistent with recent neuroscience data comparing abilities of monkeys and humans~\citep{dehaene2022symbols}. This bottleneck may also enforce a constraint on the type of joint distribution that our conscious mind can handle: the direct causal or statistical dependencies only involve a few (at most a handful) of random variables at a time~\citep{bengio2017consciousness,goyal2022inductive}. We see that in natural language, where each phrase typically only involves one verb, one subject, and a few other elements. We also see that in the mathematical formalization of science, with equations that each only involve a small number of variables.

In addition to being a bottleneck in the sense of the number of items being considered, there are reasons to believe that working memory is also a bottleneck in the sense of converting a high-dimensional continuous space (of neural activity across a large part of the brain) into a compositional discrete space (maybe corresponding to just a few words if we were to verbalize a thought). This view is supported by numerous pieces of neuroscience evidence as well as by a mathematical argument~\citep{ji2023sources} regarding contractive dynamics associated with conscious thoughts (that make it through working memory and then short-term memory). Of course, formal mathematical reasoning is manipulation of discrete symbols, constrained by a chosen set of rules. Mathematics handles continuous-valued objects by introducing abstractions for the totality of possible values (e.g., topological spaces, the field of reals), allowing particular instantiations (e.g., the number $\pi$) to be assigned symbols that can then be manipulated.

The knowledge of human cognition suggests that the brain constructs each discrete compositional thought through a generative process conditioned by the high-dimensional sensory information and past memory content. That thought goes to short-term memory and may have a chance to be retained in long-term memory. Such a process is similar to how current LLMs work, except that the symbols generated by LLMs are used directly as output, whereas our conscious thoughts are intermediate quantities, which may be better understood as unobserved (called latent) random variables. LLMs are also lacking high-dimensional sensory inputs (although multimodal generative models like GPT-4~\citep{gpt4} can also have images as inputs) and only have a memory of a finite number of past interactions, whereas humans have a recursively defined internal state that can potentially carry information from moments that are arbitrarily far in the past.

One may imagine a neural network that generates actions and thoughts as the core element providing intuitive abilities, especially when thoughts are directly turned into actions. However, we can instead deliberate in our mind -- have more thoughts -- before producing an actual output. The attention mechanism which participates in thought selection may contribute to the reasoning abilities by focusing on incoherence between the generated thoughts and previous thoughts or beliefs. In addition, one may wonder what kind of objective function may give rise to thoughts that we find interesting, and even to Eureka moments when an important mathematical idea arises. What makes a mathematical idea feel beautiful, interesting, worth communicating to others?

\section{Compression as a General Learning Theory Principle}

In an attempt to answer the above question, let us first recap potentially relevant notions from learning theory, from an information-theoretical angle. Very early results~\citep{vapnik1982estimation} about the generalization ability of learning machines point to a compression ratio: the ratio between the number of raw bits in the observed data to the number of bits to describe what has been learned. If the learner can compress the data efficiently, i.e., explain it with fewer bits, then it will provably generalize better. A similar notion emerges from the Bayesian view of statistical learning, which posits a probability distribution over models: models that do a good job at explaining the data (high likelihood of the data given as evidence) and have a small description length (high prior probability) will be given a larger weight in the posterior. Miller's `7 $\pm$ 2' estimate of working memory size~\citep{miller1956magical} and Dehaene's neuroscience observations mentioned earlier~\citep{dehaene2022symbols} are also consistent with a role of working memory as the output of a compression machinery: we remember more raw observations if they are compressible, for example because of repetitions of a pattern.
This notion of compression is also at the heart of the principle of Occam's Razor, which often guides scientific research and has inspired many machine learning approaches~\citep{rissanen1990coding,schmidhuber1992learning}: for equal ability to explain, prefer the simpler theory. The fundamental principle of Bayesian inference \citep{mackay2003information,howson2006scientific} tells us that we should prefer a theory $A$ that requires $n$ fewer bits to express than a theory $B$ by a factor of $2^n$, if both theories explain the data equally well (bits are calculated as negative prior log-probability). This exponential advantage is why good theories (that are short and fit the data) often end up dominating the other possible explanations in terms of generalization power, as ascertained by results in learning theory~\citep{vapnik1982estimation}.

These notions of compression make sense when we consider machine learning models and scientific theories, which try to explain observed data. How about mathematics, which seems detached from actual observations in the real world?

\section{Generalizing in the Space of Provable Statements}

A possible way to think about mathematics may connect the above notions of compression with the realm of theorems and proofs. But first, we need to introduce a notion of generalization in the space of provable statements, making a difference between those that are already known and proven and those that may or may not be true and provable. We are specifically interested in the exploration and discovery of new and interesting theorems, with learning agents that operate within a logical system and axioms determined by humans.

Consider the highly structured set $\mathcal{M}$ of mathematically provable statements expressed in some formal language $\mathcal{L}$, given some small set of axioms. The set $\mathcal{M}$, which contains only the provable statements, is a subset of $\mathcal{L}$, the set of all expressible statements.\footnote{Classically, in a \emph{complete} logical theory, for any statement $t$, either $t$ or its negation is provable, so $t\mapsto\neg t$ is a bijection between ${\mathcal{M}}$ and ${\mathcal{L}}\setminus{\mathcal{M}}$, modulo double negation. Thus $\mathcal{L}$ can be thought of as forming `half' of $\mathcal{M}$. However, G\"odel's incompleteness theorem shows that many interesting theories are \emph{incomplete} -- some statements cannot be proved or disproved -- and in those cases we should think of $\mathcal{M}$ as being much `smaller' than $\mathcal{L}$.} Consider similarly $\mathcal{R}$, the set of all the representable inputs to a learning machine, and its (tiny) subset $\mathcal{O}$, the set of all inputs that could actually be observed in the real world, because of physical constraints.

We suggest a parallel between the relationship of $\mathcal{O}$ to $\mathcal{R}$ and that of $\mathcal{M}$ to $\mathcal{L}$. This parallel recalls the `semantic theory of truth' \cite{tarski1943semantic,davidson1967truth} in the philosophy of language, in which the truth of a statement is identified with the satisfaction, or satisfiability, of conditions in the real world. This notion inspired the development of model theory, the branch of mathematical logic that formalizes abstraction of structure. However, here we are concerned with a different kind of generalization -- the ability to guess the truth of a statement without seeing its proof or disproof.

In the natural science / observational case, the learner only gets to see a dataset $D \subset \mathcal{O}$ of instances of all the physically possible observations. From that subset, the learner may be trained so that it would generalize outside of $D$ and estimate the probability that any $x \in \mathcal{R}$ belongs to $\mathcal{O}$. To make this more concrete, this is similar to the setting of LLMs when a given $x$ is a whole paragraph. The set of all grammatically and semantically correct paragraphs may play the role of $\mathcal{O}$, here. During training, the LLM sees a set $D$ of correct paragraphs from $\mathcal{O}$. Note that $D$ is much smaller than $\mathcal{O}$ and $\mathcal{O}$ is exponentially smaller (with respect to the length of paragraphs) than the set of all paragraph-length sequences of characters, $\mathcal{R}$. As a result of training the LLM, we actually get an estimator of $P(x \in \mathcal{O})$, the probability that a given sequence of characters $x$ could have been generated by a human following the usual rules, semantics and statistics of language\footnote{We are brushing aside the fact that maximum-likelihood training of LLMs actually does something slightly different, in that it regresses to the relative frequencies of paragraphs in a huge but finite corpus of observations of human texts. Although long paragraphs are unlikely to be repeated twice in any reasonable corpus, the smoothness of the LLM function class ensures that unseen but realistic paragraphs are still judged to have a high approximate frequency. Thus, we do not lose much in stating that the fundamental question the LLM answers is whether a given sequence of characters is language-like or not.}.

\section{Proofs as Trajectories for Conjecture-Making Agents}

The above analogies introduce a notion of generalization in the ability to guess that a given yet unseen element (statement or observation) is valid (true or observable) but it misses the notion of proof, which connects true statements together to form trajectories and graphs describing ways to prove a theorem. Let us try to do that and relate it to the concepts involved in reinforcement learning~\citep{sutton2018reinforcement} or RL, with agents that trace out trajectories in a state-space.

An agent acting within an environment, as formalized by the notion of Markov decision process~\citep{sutton2018reinforcement}, is at any point in a state $S$ and takes an action $A$, which brings it a reward $R$ and into a new state $S'$, deterministically or stochastically. The actions of a conjecture-making agent consists in a derivation step, in which previously known theorems and axioms are used to derive a new statement. Standard RL optimizes the policy (e.g. a neural network) that chooses actions in each state so as to maximize the expected value of future rewards.

In mathematics, the derivation rules are transformations known as \emph{proof tactics}~\citep{milner1984use}; modern proof verification systems, such as Coq~\citep{coq} and Lean~\citep{moura2021lean} implement thousands of predefined tactics.
Such an action thus modifies a directed acyclic graph where each node is a true statement and the edges into a node indicate how to derive the corresponding statement mechanically from its parents.
For our case of interest with conjecture-making agents, we can thus define the state $S$ to be such a graph. Given a body of known theorems specified by $S$, a proof is thus a directed and connected subgraph of statements derived from $S$ and from each other and leading into a new node for the proven theorem.

There is an interesting subtlety we need to add to the above notions. In order to prove a new theorem, we only need to refer to previous theorems, not to the intermediate statements that belong to earlier proofs (although the abstract ideas to orient those proofs and make them efficient may be exploited by a skilled mathematician by being part of their experience and analogy-formation abilities\footnote{``A mathematician is a person who can find analogies between theorems; a better mathematician is one who can see analogies between proofs and the best mathematician can notice analogies between theories. One can imagine that the ultimate mathematician is one who can see analogies between analogies." (attributed to Stefan Banach in \citet{banachspaces}).}). In other words, the new proof only needs to refer to a subset $T(S)$ of $S$, the subset of theorems, i.e., excluding the intermediate statements made in their proofs. A state of our conjecture-making agent thus consists in a directed graph $S$ where some nodes (those in $T(S)$) have a special label: they are not just proven statements, they are theorems, i.e., they are the basis from which new theorems are to be proven.

The working memory bottleneck of human brains, in this context, would mean that each derivation should only involve the combination of a very small number of antecedent statements.
It would be interesting to test whether such constraints are experimentally advantageous. The constraint could become more interesting in the context where the learner can choose to define new symbols and tactics that can be reused in the same proof or even more globally, since this seems to be a specifically human aptitude~\citep{dehaene2022symbols}.

\section{How Useful is a Theorem?}

That leaves us with the next question: what is it that could motivate such a conjecture-making agent? What is it that makes a next state $S'$ better than another?

Clearly, a theorem's usefulness depends on the context of the body of previously known theorems, i.e., forming a new state $S'$ by grafting the proof subgraph $p$ for $t$ onto some earlier state $S$.

Before talking about theorems, consider what a generative model of images does: it lossily compresses the infinite set of all plausible natural images into a generative model with a finite number of bits (with a finite number of finite-precision parameters).

By analogy, we can now state the main hypothesis proposed in this paper: {\bf a crucial component of the usefulness of a new proven theorem $t$ (in the context of previous theorems $T(S)$) is how efficiently $T(S)\cup \{t\}$ compresses the set of all provable mathematical statements $\mathcal{M}$.} That is, $T(S)\cup\{t\}$ is a good compression of $\mathcal{M}$ if many provable statements, beyond those in $S$, can be derived from $T(S)\cup\{t\}$, say using at most $k$ derivation steps.

The compression efficiency of $T(S)$ depends both on its ability to prove new theorems in a few steps and on the size of $T(S)$ in number of bits of description. For example, a single theorem that is the logical conjunction of a large number of (sub-)theorems would have about the same description length as the set of these theorems taken as a mathematical theory.

A related metric is how well $S$, as training data, enables a learner to generalize correctly on the provability of statements $s' \in {\mathcal{M}}\setminus S$, {\em before doing the proof} of $s'$, i.e., based on a kind of machine learning intuition obtained thanks to the `data' formed by $S$. The latter is much easier to estimate if we are given a distribution over statements (that can be provably true or not) and we could check how accurately the learner manages to predict their provability before even trying to prove them. A held-out subset of mathlib~\citep{mathlib2019} could be used to quantitatively assess such a predictive ability, for example. Alternatively, one could consider a set of statements selected for didactic value or estimated difficulty for human mathematicians, as has been done from the earliest geometry problem solvers that predate modern machine learning~\citep{gelernter1959realization} to modern efforts to solve mathematical Olympiad problems with AI~\citep{zheng2021mini,imogc}.

The above ideas would make an AI mathematician more like an AI scientist, where experiments are like proofs and the outcome of experiments like theorems: a good scientific theory allows to predict the outcome of an experiment without having to actually perform the experiment. Similarly, a good understanding of mathematics derived from a set of theorems and their proofs may enable a learner to correctly guess the provability of many new theorems. In this analogy we want both the chosen body of theorems and the chosen scientific theory to be compact in terms of their description length, according to Occam's Razor.

What we would like to end up with is an objective function for training an AI mathematician that does not require human labels of interestingness (although pre-training on existing theorems that humans find interesting is probably a good idea from the point of view of efficiently training the learner). Note that if $t'$ is an ugly variant of $t \in T(S)$ that does not really help proving other theorems better than $t$ does, and if $t'$ is not shorter than $t$ in its description length, then there is no advantage according to our above proposals to include $t'$ instead of $t$ in $T(S)$ (and even less advantage to include $t'$ in addition to $t$).

In practice, the development of mathematics has often been motivated by applications in all spheres of human activity, from simple enumeration and measurement to the study of the most complex natural phenomena. These applications give us other reasons for liking some theorems. For example, in computer science, theorems may help us ascertain the usefulness of an algorithm or guarantees about its runtime. We will leave to future work such investigations of how to quantify additional indicators of theorems in the context of their applications. However, many theorems that were discovered because they felt interesting and beautiful later turned out to also be useful in applications -- a remarkable testament to the value of pure mathematics. Thus, studying the intrinsic `interestingness' of theorems seems a worthy endeavor.

\section{Active Learning}

The generalization ability of a learner, as a criterion for the usefulness of its training set, is an old concept in machine learning, arising from the idea of active learning~\citep{settles2009active}. Whereas the most widely studied paradigm of statistical learning theory considers the training data of a learner to be fixed and given, in many real-life scenarios, for example with learning agents that interact with their environment~\citep{sutton2018reinforcement}, the learner can make some choices that influence what new data will be added to its training set. The setting in which a learner can influence the choices of its training data is known as active learning, and the exploration problem~\citep{amin2021survey} concerns the agent's choice of what new data to add to its training set. The classical application of such active learning is in deciding which unlabeled example images $x$ should be sent to human labelers to be associated with a target output $y$. For example, when creating an image classifier, one may collect hundreds of millions of images from the Internet but carefully select only some of them to be categorized by humans, reducing the cost of the expensive labeling process for a given final level of error on new unlabeled images. This will also generally be an iterative process that can greatly accelerate learning, in the sense that fewer examples are needed to achieve the same generalization accuracy or better generalization is obtained with the same amount of training examples. In the image labeling scenario, depending on which images have already been labeled, the current learner may have more or less uncertainty -- called epistemic uncertainty~\citep{hullermeier2021aleatoric} -- regarding the correct label of unlabeled images. A good heuristic is to select the images on which the current iteration of the classifier is highly uncertain. These new (image,label) pairs are added to the training set for the next iteration of fitting the classifier, which will generally improve the ability of the classifier to generalize accurately on new images. Interestingly, active learning makes the generalization ability provably converge at faster rates~\citep{hanneke2011rates} than passive learning (for which the examples are picked in a fixed random way from some distribution).

An extreme example illustrating why active learning can be much faster is the hi-lo game, where it can be exponentially faster. Consider a one-dimensional threshold-based classifier on the set of integers from 1 to $N$. The true classifier is $y=f(x)=1_{x>\theta}$ but $\theta$ is unknown and should be estimated. A passive learner would just collect $n$ uniform samples from $[1,N]$ and construct a new estimate $\hat{\theta}$ by finding the median between the largest of the $y=0$ examples and the smallest of the $y=1$ examples. An active learner would use its budget of $n$ queries sequentially and smartly: at each step, it would ask to label $x$ whose value is its current estimate $\hat{\theta}$ for $\theta$, since this is the most uncertain value of $x$. At each step, the interval where the true $\theta$ must lie is at least halved (depending on whether the query is successful or not, we pick either the left of right subinterval split by $\hat{\theta}$), which guarantees exponential convergence. This is the optimal strategy in the hi-lo game and it requires exponentially less examples (with respect to $N$) to converge to the same expected error as the passive learner, and this can be shown in other settings as well~\citep{cohn1994improving}.

In our conjecture-making context, the epistemic uncertainty heuristic (pick examples that are less certain under the current learner's model) may suggest that {\bf useful theorems are also surprising theorems}. Ex-post, after seeing the proof of an unexpectedly true theorem, i.e., if $P(t | T(S))$ is small according to the "intuitive" prediction of the AI mathematician, it would maybe be useful to add $t$ to $T(S)$. Ex-ante, i.e., before we see a proof (or a proof that the statement is false), we may rely on the estimated entropy $\mathbb{E}_{P(t|T(S))}[-\log P(t\mid T(S))]$ of the truth value of the statement $t$ to ascertain the information gain, or expected surprise, i.e., how useful it may be to search for a proof or a counterexample of $t$, given the current body of theorems.

Active learning is related to another interesting notion from machine learning that relates to mathematics education: curriculum learning~\citep{bengio2009curriculum}. The idea is that instead of being presented examples in a random order, the learner is shown examples in an order that depends on the learner's current state of understanding, typically with easier and simpler concepts being illustrated before more complex abstractions. Again, we see that `interestingness' of a new theorem depends on the context of the learner: not just what the learner has already seen but also the learner's strengths and weaknesses, how well they have digested different concepts. In active learning, the ability of the learner to make good predictions or their uncertainty regarding the correct answer for new questions plays a crucial role. Curriculum learning has successfully been used in conjunction with the most advanced automated theorem proving systems based on large language models~\citep{polu2023formal}.

\section{Conjecture Generation as Goal-Conditioned Exploration}

Armed with some indicator $R(t)$ of the interestingness of any given conjecture $t$, how do we generate interesting theorems, i.e., with a high value of $R(t)$? We can't possibly screen all the possible statements $t$ and since the set of provable statements is a tiny subset of all statements, random sampling of statements is not going to work either.

Another option is to use a generative flow network or GFlowNet~\citep{bengio2021gflownet}. A GFlowNet is a generative model of structured objects (such as a theorem, seen either as a string or a more compositionally structured object, like a program and its parse tree). Unlike standard generative models, it can be trained not just from positive examples (of what is good to generate) but also or even only from a queryable reward function $R(\cdot)$ that characterizes the usefulness of its argument. The global minimum of its training objective will make it sample objects $t$ with probability proportional to $R(t)$. This is also different from reinforcement learning methods, which search for a single $t$ that maximizes the reward $R(t)$. The advantage of the GFlowNet is that it is by nature more exploratory: it is looking for all the interesting theorems. Once a GFlowNet is trained, we can efficiently sample from it a bunch of independently sampled conjectures, which may thus cover a more diverse set of interesting conjectures.

Having potentially interesting statements as conjectures in hand, we still need to try to prove them.
In the reinforcement learning jargon, we use the term \emph{goal-conditioned policy} to talk about a generator of sequences of actions (here to prove a theorem) that is provided a goal as a target to reach.
In the practice of discovering theorems, however, humans need not start from the correct theorem statement. In searching for a proof for a conjecture, they may modify the goal by suggesting alternative formulations (e.g., by adding additional assumptions) that are easier to prove (or just provable, while the original goal was unreachable). On the other hand, the goal may ultimately be replaced with a result far more general than the original statement: examples abound of conjectures with simple statements spurring the development of entire branches of mathematics (sometimes over the course of centuries, as Fermat's Last Theorem led to modern algebraic number theory).
How to formalize this task of modifying the goal is an open question even in the broader reinforcement learning literature.

\section{Proof Plans and Lemmas as Subgoals}

What kind of machine learning algorithms could help in the subtask of finding a proof, given a conjecture as a goal?
Some of the most successful existing theorem-proving algorithms are large language models trained by \emph{imitation learning} -- learning to mimic human-written proofs~\citep{han2022proof}. However such models never get the chance to learn from theorems for which no proof is provided (like a proof given as homework or left as the proverbial `exercise to the reader').
On the other hand, \emph{goal-conditioned reinforcement learning} algorithms are similar to ordinary deep reinforcement learning methods except that the neural network receives an extra input that specifies the goal, and the reward (to be maximized) also depends on that goal. There are already several such methods~\citep{crouse2021deep} that have been proposed and should serve as a starting point. We remark that producing a valid proof using a reinforcement learning policy may still require an extensive search, in which the trained prover suggests trajectories of tactics until a successful one is found.

However, human mathematicians find proofs in a way that could serve as additional inspiration for the AI system. Humans come up with conjectured lemmas as subgoals, as waypoints on the path to proving the target theorem. There is also a rich literature on the topic of planning, hierarchical reinforcement learning and subgoal generation~\citep{kim2021landmark} which could serve as inspiration here.

A particularly interesting aspect of mathematics which is relevant to our starting compression viewpoint is that mathematicians give names to theorems, lemmas and rewrite tactics, as well as define new variables (e.g., that satisfy given equations). All these new names introduced along the way can make downstream proofs and theorem statements more compact and sometimes easier to understand (especially if these names correspond to meaningful abstractions that human minds can latch onto). Training objectives for the AI mathematicians which have an explicit notion of compression would therefore naturally attempt to create intermediate definitions. If a new symbol is used more than once, then its definition is already useful from the point of view of compressing the overall sequence of statements. The principle of compression has been used to propose new definitions by identifying repeating parts in a body of \emph{existing} proofs \citep{vyskocil} and is also used in existing AI systems for program synthesis~\citep{ellis2021dreamcoder}, a close cousin of theorem proving\footnote{According to the Curry-Howard correspondence~\citep{howard1969formulae,wadler2015propositions} and modern type theory interpretation of proofs, theorems are to be seen as types (akin to classes in object-oriented programming), while proofs are compilable programs that evaluate to elements of these types, serving as evidence of their provability.}, while retrieval-based systems take advantage of the generalizability of meaningful, human-readable names for statements and tactics~\citep{yang2023leandojo}.

Another interesting aspect of naming of mathematical entities is that some names are local, only useful within a proof (e.g., a lemma) or in order to state a theorem (e.g., defining assumptions, symbols, etc), while other names (e.g., of theorems) have a global scope. This sheds light on our above proposal to only use theorems (and not the statements in their proofs) as starting points to prove new theorems. It not only means that a more compact representation of the acquired knowledge is possible (including only the theorems) but it may also help a conjecture-making and theorem-proving learner by reducing the set of statements it needs to consider to form any derivation step.

Generating several candidate paths and estimating intermediate states in the search process is a common staple of deep reinforcement learning systems, such as the famous AlphaGo~\citep{silver2017mastering}, which uses a variant of the Monte Carlo Tree Search approach to efficiently prune potential paths in the search space. This is clearly a feature of deliberate reasoning abilities, although human minds seem able to find good paths with very few actual candidate paths being considered, in comparison with AlphaGo.
Another relevant feature of AlphaGo is that unlike its predecessors, it did not need to be trained or even pre-trained by imitating very strong human players: it could fully learn to play Go solely through self-play.
Human mathematicians benefit a lot from studying the existing proofs and theorems, so it may be reasonable to try to combine both pre-training from a corpus like mathlib~\citep{mathlib2019} and downstream unsupervised exploration to discover potentially new and improved ways of playing the game of mathematics.

\section{Conclusion}

In this paper, we only outlined a set of ideas towards not just proving theorems but also, and maybe more interestingly, discovering them, leaving many open questions along the way.

The objective of compressing all provable statements is too expensive to compute directly, but approximations and derived heuristics may be good first steps.
An interesting general principle that has worked incredibly well for scaling deep learning to extremely complex tasks is that of stochastic gradient-based optimization. We do not need to compute the training objective of interest: it suffices to compute an estimator of its gradient whose expected value is correct (and even a small bias may still be acceptable). For example, it may be sufficient to look at how and how often a theorem is used to get a sense of how useful it may be in terms of compressing the set of provable statements.

The active learning and reinforcement learning literature, especially the unsupervised or purely exploratory reinforcement learning methods, should also be good sources of inspiration.
Conversely, making progress on designing an AI mathematician could help progress in other areas of AI research, in particular in AI for science applications, where we would like the learner to discover a compact representation of a set of facts by coming up with new and useful abstractions.

\newpage

\bibliographystyle{plainnat}
\bibliography{references}

\end{document}